%% file: emnlp2017.tex
\pdfoutput=1
\documentclass[11pt,letterpaper,table]{article}
\usepackage{emnlp2017}
\PassOptionsToPackage{hyphens}{url}
\usepackage{hyperref}
\usepackage{times}
\usepackage{latexsym}

\emnlpfinalcopy

\author{
Johannes Daxenberger\textsuperscript{\textdagger}, 
Steffen Eger\textsuperscript{\textdagger\textdaggerdbl}, 
Ivan Habernal\textsuperscript{\textdagger},
Christian Stab\textsuperscript{\textdagger}, 
Iryna Gurevych\textsuperscript{\textdagger\textdaggerdbl}\\
\textsuperscript{\textdagger}Ubiquitous Knowledge Processing Lab (UKP-TUDA)\\
Department of Computer Science, Technische Universit{\"a}t Darmstadt\\
\textsuperscript{\textdaggerdbl}Ubiquitous Knowledge Processing Lab (UKP-DIPF)\\
German Institute for Educational Research and Educational Information\\
\url{http://www.ukp.tu-darmstadt.de}
}

\usepackage[utf8]{inputenc}
\usepackage{amsmath,amssymb}
\usepackage{tabularx}
\usepackage{multirow}
\usepackage{booktabs}
\usepackage{comment}
\usepackage{mathptmx}
\usepackage{graphicx}
\usepackage{bbm}
\usepackage{pgfplotstable}
\pgfplotsset{compat=1.13}

\usepackage{etoolbox}
\appto\UrlBreaks{\do\a\do\b\do\c\do\d\do\e\do\f\do\g\do\h\do\i\do\j
\do\k\do\l\do\m\do\n\do\o\do\p\do\q\do\r\do\s\do\t\do\u\do\v\do\w
\do\x\do\y\do\z}
\expandafter\def\expandafter\UrlBreaks\expandafter{\UrlBreaks
  \do\a\do\b\do\c\do\d\do\e\do\f\do\g\do\h\do\i\do\j%
  \do\k\do\l\do\m\do\n\do\o\do\p\do\q\do\r\do\s\do\t%
  \do\u\do\v\do\w\do\x\do\y\do\z\do\A\do\B\do\C\do\D%
  \do\E\do\F\do\G\do\H\do\I\do\J\do\K\do\L\do\M\do\N%
  \do\O\do\P\do\Q\do\R\do\S\do\T\do\U\do\V\do\W\do\X%
  \do\Y\do\Z\do\*\do\-\do\~\do\'\do\"\do\-}%
\pgfplotstableset{
    colorcells/.style={
        col sep=comma,
        string type,
        postproc cell content/.code={%
                \pgfkeysalso{@cell content=\rule{0cm}{2.4ex}\cellcolor{black!##1}\pgfmathtruncatemacro\number{##1}\ifnum\number>65\color{white}\fi##1}%
                },
        columns/x/.style={
            column name={},
            postproc cell content/.code={}
        }
    }
}

\title{What is the Essence of a Claim? Cross-Domain Claim Identification}

\date{}

\begin{document}

\maketitle

\begin{abstract}
Argument mining has become a popular research area in NLP. 
It typically includes the identification of argumentative components, e.g.\ claims, as the central component of an argument.
We perform a qualitative analysis across six different datasets and show that these appear to conceptualize claims quite differently. 
To learn about the consequences of such different conceptualizations of claim for practical applications, we carried out extensive experiments using state-of-the-art feature-rich and deep learning systems, to identify claims in a cross-domain fashion.
While the divergent conceptualization of claims in different datasets is indeed harmful to cross-domain classification, we show that there are shared properties on the lexical level as well as system configurations that can help to overcome these gaps.
\end{abstract}

\section{Introduction}
\label{sec:intro}

The key component of an argument is the \emph{claim}. 
This simple observation has not changed much since the early works on argumentation by Aristotle more than two thousand years ago, although argumentation scholars provide us with a plethora of often clashing theories and models \cite{vanEemeren.et.al.2014}.
Despite the lack of a precise definition in the contemporary argumentation theory,
Toulmin's influential work on argumentation in the 1950's introduced a claim as an `assertion that deserves our attention' \cite[p.~11]{Toulmin.2003}; recent works describe a claim as `a statement that is in dispute and that we are trying to support with reasons' \cite{Govier.2010}.

Argument mining, a computational counterpart of manual argumentation analysis, is a recent growing sub-field of NLP \cite{Peldszus2013}.
`Mining' arguments usually involves several steps like separating argumentative from non-argumentative text units, parsing argument structures, and recognizing argument components such as claims---the main focus of this article.
Claim identification itself is an important prerequisite for applications 
such as fact checking \cite{vlachos-riedel:2014:W14-25}, politics and legal affairs \cite{surdeanu-nallapati-manning:2010:HierarchicalIE}, and science \cite{Park:2012:ICC:2391171.2391173}.

Although claims can be identified with a promising level of accuracy in typical argumentative discourse such as persuasive essays \cite{Stab2014b,Eger2017}, less homogeneous resources, for instance online discourse, pose challenges to current systems \cite{Habernal2016}. 
Furthermore, existing argument mining approaches are often limited to a single, specific domain like legal documents \cite{MochalesPalau2009}, microtexts \cite{Peldszus2015}, Wikipedia articles \cite{Levy2014,Rinott2015} or student essays \cite{Stab2016}.
The problem of generalizing systems or features and their robustness across heterogeneous datasets thus remains fairly unexplored.

This situation motivated us to perform a detailed analysis of the concept of claims (as a key component of an argument) in existing argument mining datasets from different domains.\footnote{We take the machine learning perspective in which different \emph{domains} mean data drawn from different distributions \cite[p.~297]{Murphy.2012}.}
We first review and qualitatively analyze six existing publicly available datasets for argument mining (§\ref{sec:data}), showing that the conceptualizations of claims in these datasets differ largely. 
In a next step, we analyze the influence of these differences for cross-domain claim identification.
We propose several computational models for claim identification, including systems using linguistically motivated features (§\ref{sec:features}) and recent deep neural networks (§\ref{sec:deep-learning}), and rigorously evaluate them on and across all datasets (§\ref{sec:results}).
Finally, in order to better understand the factors influencing the performance in a cross-domain scenario, we perform an extensive quantitative analysis on the results (§\ref{sec:analysis}).

Our analysis reveals that despite obvious differences in conceptualizations of claims across datasets, there are some shared properties on the lexical level which can be useful for claim identification in heterogeneous or unknown domains.
Furthermore, we found that the choice of the source (training) domain is crucial when the target domain is unknown.
We release our experimental framework to help other researchers build upon our findings.\footnote{\url{https://github.com/UKPLab/emnlp2017-claim-identification}}

\section{Related Work}
\label{sec:relatedWork}

Existing approaches to argument mining can be roughly categorized into (\emph{a}) \emph{multi-document} approaches which recognize claims and evidence across several documents and (\emph{b}) \emph{discourse level} approaches addressing the argumentative structure within a single document.
Multi-document approaches have been proposed e.g.\ by \newcite{Levy2014} and \newcite{Rinott2015} for mining claims and corresponding evidence for a predefined topic over multiple Wikipedia articles. 
Nevertheless, to date most approaches and datasets deal with single-document argumentative discourse. 
This paper takes the discourse level perspective, as we aim to assess multiple datasets from different authors and compare their notion of `claims'.

\newcite{MochalesPalau2009} experiment at the discourse level using feature-rich SVM and a hand-crafted context-free grammar in order to recognize claims and premises in legal decisions.
Their best results for claims achieve $74.1\%$ $F_1$ using domain-dependent key phrases, token counts, location features, information about verbs, and the tense of the sentence. 
\newcite{Peldszus2015} present an approach based on a minimum spanning tree algorithm and model the global structure of arguments considering argumentative relations, the stance and the function of argument components. 
Their approach yields  $86.9\%$ $F_1$ for recognizing claims in English `microtexts'.
\newcite{Habernal2016} cast argument component identification as BIO sequence labeling and jointly model separation of argumentative from non-argumentative text units and identification of argument component boundaries together with their types.
They achieved $25.1\%$ Macro-$F_1$ with a combination of topic, sentiment, semantic, discourse and embedding features using structural SVM.
\newcite{Stab2014b} identified claims and other argument components in student essays. 
They experiment with several classifiers and achieved the best performance of $53.8\%$ $F_1$ score using SVM with structural, lexical, syntactic, indicator and contextual features.
Although the above-mentioned approaches achieve promising results in particular domains, their ability to generalize over heterogeneous text types and domains remains unanswered.

\newcite{Rosenthal:2012} set out to explore this direction by conducting cross-domain experiments for detecting claims in blog articles from LiveJournal and discussions taken from Wikipedia. 
However, they focused on relatively similar datasets that both stem from the social media domain and in addition annotated the datasets themselves, leading to an identical conceptualization of the notion of claim.
Although \newcite{stein:2016} also deal with cross-domain experiments, they address a different task; namely identification of argumentative sentences.
Further, their goals are different: they want to improve argumentation mining via distant supervision rather than detecting differences in the notions of a claim.

\begin{table*}[t]
\footnotesize{
	\begin{tabularx}{\textwidth}{ p{0.05\textwidth} p{0.25\textwidth}  p{0.14\textwidth} p{0.06\textwidth}  p{0.1\textwidth} p{0.08\textwidth} p{0.13\textwidth} }
\toprule
	 \centering{\emph{Corpus}} & \centering{\emph{Reference}} & \centering{\emph{Genre}} & \centering{\emph{\#Docs}} & \centering{\emph{\#Tokens}} & \centering{\emph{\#Sentences}} & \centering{\emph{\#Claims}} \tabularnewline
\midrule
\centering{\textbf{VG}} & \centering{\newcite{Reed2008}} & \centering{various genres} & \centering{$507$} & \centering{$60{,}383$} & \centering{$2{,}842$} & \centering{$563$ ($19.81\%$)} \tabularnewline

\centering{\textbf{WD}} & \centering{\newcite{Habernal2015}} & \centering{web discourse} & \centering{$340$} & \centering{$84{,}817$} & \centering{$3{,}899$} & \centering{$211$ ($5.41\%$)} \tabularnewline

\centering{\textbf{PE}} & \centering{\newcite{Stab2016}} & \centering{persuasive essays} & \centering{402} & \centering{$147{,}271$} & \centering{$7{,}116$} & \centering{$2{,}108$ ($29.62\%$)} \tabularnewline

\centering{\textbf{OC}} & \centering{\newcite{Biran2011a}} & \centering{online comments} & \centering{$2{,}805$} & \centering{$125{,}677$} & \centering{$8{,}946$} & \centering{$703$ ($7.86\%$)}\tabularnewline

\centering{\textbf{WTP}} & \centering{\newcite{Biran2011b}} & \centering{wiki talk pages} & \centering{$1{,}985$} & \centering{$189{,}140$} & \centering{{9{,}140}} & \centering{$1{,}138$ ($12.45\%$)}\tabularnewline

\centering{\textbf{MT}} & \centering{\newcite{Peldszus2015}} & \centering{micro texts} & \centering{112} & \centering{$8{,}865$} & \centering{$449$} & \centering{$112$ ($24.94\%$)} \tabularnewline

\bottomrule
	\end{tabularx}	
	\caption{Overview of the employed corpora.}\label{tab:corpora}
}
\end{table*} 

Domain adaptation techniques \cite{Daume:2007} try to address the frequently observed drop in classifier performances entailed by a dissimilarity of training and test data distributions.
Since techniques such as learning generalized cross-domain representations  
in an unsupervised manner 
\cite{Blitzer:2006,Pan:2010,Glorot:2011,Yang:2015}
have been criticized for targeting specific source and target domains,
it has alternatively been proposed 
to learn \emph{universal} representations from general domains in order to render a learner robust across \emph{all} possible domain shifts \cite{Mueller:2015,Schnabel:2013}.  
Our approach is in a similar vein. 
However, rather than trying to improve classifier performances for a specific source-target domain pair, we want to detect \emph{differences} between these pairs. 
Furthermore, we are looking for \emph{universal} feature sets or classifiers that perform generally well for claim identification across varying source and target domains. 

\section{Claim Identification in Computational Argumentation}
\label{sec:data}

We briefly describe six English datasets used in our empirical study; they all capture claims on the discourse level.
Table \ref{tab:corpora} summarizes the dataset statistics relevant to claim identification.

\subsection{Datasets}

The AraucariaDB corpus \cite{Reed2008} includes various genres (\textbf{VG}) such as newspaper editorials, parliamentary records, or judicial summaries. 
The annotation scheme structures arguments as trees and distinguishes between claims and premises at the clause level.
Although the reliability of the annotations is unknown, the corpus has been extensively used in argument mining \cite{Moens2007,Feng2011,Rooney2012}. 

The corpus from \newcite{Habernal2016} includes user-generated web discourse (\textbf{WD}) such as blog posts, or user comments annotated with claims and premises as well as backings, rebuttals and refutations ($\alpha_U\ 0.48$) inspired by Toulmin's model of argument \cite{Toulmin.2003}.


The persuasive essay (\textbf{PE}) corpus \cite{Stab2016} includes $402$ student essays.
The scheme comprises major claims, claims and premises at the clause level ($\alpha_U\ 0.77$).
The corpus has been extensively used in the argument mining community \cite{Persing2015,Lippi2015,Nguyen2016}.


\newcite{Biran2011a} annotated claims and premises in online comments (\textbf{OC}) from blog threads of LiveJournal ($\kappa\ 0.69$).
In a subsequent work, \newcite{Biran2011b} applied their annotation scheme to documents from  Wikipedia talk pages (\textbf{WTP}) and annotated $118$ threads. 
For our experiments, we consider each user comment in both corpora as a document, which yields $2,805$ documents in the OC corpus and $1,985$ documents in the WTP corpus.


\newcite{Peldszus2016} created a corpus of German microtexts (\textbf{MT}) of controlled linguistic and rhetoric complexity. 
Each document includes a single argument and does not exceed five argument components. 
The scheme models the argument structure and distinguishes between premises and claims, among other properties (such as proponent/opponent or normal/example).
In the first annotation study, $26$ untrained annotators annotated $23$ microtexts in a classroom experiment ($\kappa\ 0.38$) \cite{Peldszus2013b}.
In a subsequent work, the corpus was largely extended by expert annotators ($\kappa\ 0.83$).
Recently, they translated the corpus to English, resulting in the first parallel corpus in computational argumentation; our experiments rely on the English version.

\subsection{Qualitative Analysis of Claims}

In order to investigate how claim annotations are tackled in the chosen corpora, one co-author of this paper manually analyzed 50 randomly sampled claims from each corpus.
The characteristics taken into account are drawn from argumentation theory \cite{Schiappa.Nordin.2013} and include among other things the claim type, signaling words and discourse markers.

\newcite{Biran2011b} do not back-up their claim annotations by any common argumentation theory but rather state that claims are \emph{utterances which convey subjective information and anticipate the question `why are you telling me that?'}\ and need to be supported by justifications. 
Using this rather loose definition, a claim might be any subjective statement that is justified by the author. 
Detailed examination of the LiveJournal corpus (OC) revealed that sentences with claims are extremely noisy. 
Their content ranges from a single word,
 (\emph{``Bastard.''}), 
 to emotional expressions of personal regret,
 (\emph{``::hugs:: i am so sorry hon ..''})
  to general Web-chat nonsense 
 (\emph{``W-wow... that's a wicked awesome picture...  looks like something from Pirates of the Caribbean...gone Victorian ...lolz.''}) 
  or posts without any clear argumentative purpose
 (\emph{``what i did with it was make this recipe for a sort of casserole/stratta (i made this up, here is the recipe) [...] butter, 4 eggs, salt, pepper, sauted onions and cabbage..add as much as you want bake for 1 hour at 350 it was seriously delicious!''}).
The Wikipedia Talk Page corpus (WTP) contains claims typical to Wikipedia quality discussions  
 (\emph{``That is why this article has NPOV issues.''}) 
and policy claims \cite{Schiappa.Nordin.2013} are present as well 
 (\emph{``I think the gallery should be got rid of altogether.''}). 
However, a small number of nonsensical claims remains
 (\emph{``A dot.''}).

Analysis of the MT dataset revealed that about half of claim sentences contain the modal verb \emph{`should'}, clearly indicating policy claims
 (\emph{``The death penalty should be abandoned everywhere.''}).
Such statements also very explicitly express the stance on the controversial topic of interest.
In a similar vein, claims in persuasive students' essays (PE) heavily rely on phrases signaling beliefs 
 (\emph{``In my opinion, although using machines have many benefits, we cannot ignore its negative effects.''}) 
or argumentative discourse connectors whose usage is recommended in textbooks on essay writing
 (\emph{``Thus, it is not need for young people to possess this ability.''}).
Most claims are value/policy claims written in the present tense.

The mixture of genres in the AraucariaDB corpus (VG) is reflected in the variety of claims. While some are simple statements starting with a discourse marker
 (\emph{``Therefore, 10\% of the students in my logic class are left-handed.''}), 
there are many legal-specific claims requiring expert knowledge 
(\emph{``In considering the intention of Parliament when passing the 1985 Act, or perhaps more properly the intention of the draftsman in settling its terms, there are [...]''}),
reported and direct speech claims
 (\emph{``Eight-month-old Kyle Mutch's tragic death was not an accident and he suffered injuries consistent with a punch or a kick, a court heard yesterday.''}),
and several nonsensical claims 
(\emph{``RE: Does the Priest Scandal Reveal the Beast?''}) 
which undercut the consistency of this dataset.

The web-discourse (WD) claims take a clear stance to the relevant controversy
 (\emph{``I regard single sex education as bad.''}),
yet sometimes anaphoric
 (\emph{``My view on the subject is no.''}).
The usage of discourse markers is seldom.
\newcite{Habernal2016} investigated hedging in claims and found out that it varies with respect to the topic being discussed (10\% up to 35\% of claims are hedged). 
Sarcasm or rhetorical question are also common
 (\emph{``In 2013, is single-sex education really the way to go?''}).

These observations make clear that annotating claims---the central part of all arguments, as suggested by the majority of argumentation scholars---can be approached very differently when it comes to actual empirical, data-driven operationalization. While some traits are shared, such as that claims usually need some support to make up a `full' argument (e.g., premises, evidence, or justifications), the exact definition of a claim can be arbitrary---depending on the domain, register, or task.

\section{Methodology}
\label{sec:approach}

Given the results from the qualitative analysis, we want to investigate whether the different conceptualizations of claims can be assessed empirically and if so, how they could be dealt with in practice.   
Put simply, the task we are trying to solve in the following is: \textit{given a sentence, classify whether or not it contains a claim.}
We opted to model the claim identification task on sentence level, as this is the only way to make all datasets compatible to each other. 
Different datasets model claim boundaries differently, e.g. MT includes discourse markers within the same sentence, whereas they are excluded in PE. 

All six datasets described in the previous section have been preprocessed by first segmenting documents into sentences using Stanford CoreNLP \cite{Manning.2014} and then annotating every sentence as claim, if one or more tokens within the sentence were labeled as claim (or major claim in PE). 
Analogously, each sentence is annotated as non-claim, if none of its tokens were labeled as claim (or major claim).
Although our basic units of interest are sentences, we keep the content of the entire document to be able to retrieve information about the context of (non-)claims.\footnote{This is true only for the feature-based learners. The neural networks do not have access to information beyond individual sentences.}

We are not interested in optimizing the properties of a certain learner for this task, but rather want to compare the influence of different types of lexical, syntactical, and other kinds of information across datasets.\footnote{For the same reason, we do not optimize any hyperparameters for individual learners, unless explicitly stated.}
Thus, we used a limited set of learners for our task:
a) a standard L2-regularized logistic regression approach with manually defined feature sets\footnote{Using the liblinear library \cite{Fan:2008}.}, which is a simple yet robust and established technique for many text classification problems \cite{Plank.et.al.2014.ACL,He.et.al.2015,Zhang.et.al.2016.NAACL,Ferreira.Vlachos.2016.NAACL}; and 
b) several deep learning approaches, using state-of-the-art neural network architectures. 

The \textbf{in-domain experiments} were carried out in a 10-fold cross-validation setup with fixed splits into training and test data. 
As for the \textbf{cross-domain experiments}, we train on the entire data of the source domain and test on the entire data of the target domain. 
In the domain adaptation terminology, this corresponds to an unsupervised setting.

To address class-imbalance in our datasets (see Table~\ref{tab:corpora}), 
we downsample the negative class (non-claim) both in-domain and cross-domain, so that positive and negative class occur approximately in an 1:1 ratio in the training data.
Since this means that we discard a lot of useful information (many negative instances), we repeat this procedure 20 times, in each case randomly discarding instances of the negative class such that the required ratio is obtained. 
At test time, we use the majority prediction of this ensemble of 20 trained models. 
With the exception of very few cases, this led to consistent performance improvements across all experiments.
The systems are described in more detail in the following subsections.
Additionally, we report the results of two baselines.
The majority baseline labels all sentences as non-claims (predominant class in all datasets), the random baseline labels sentences as claims with 0.5 probability.

\subsection{Linguistically Motivated Features}
\label{sec:features}

For the logistic regression-based experiments (LR) we employed the following feature groups. 
\emph{Structure Features} capture the position, the length and the punctuation of a sentence.
\emph{Lexical Features} are lowercased unigrams.
\emph{Syntax Features} account for grammatical information at the sentence level. We include information about the part-of-speech and parse tree for each sentence.
\emph{Discourse Features} encode information extracted with help of the Penn Discourse Treebank (PDTB) styled end-to-end discourse parser as presented by \newcite{Lin2014}.
\emph{Embedding Features} represent each sentence as a summation of its word embeddings \cite{Guo.et.al.2014}. 
We further experimented with sentiment features \cite{Habernal2015,Anand2011a} and dictionary features \cite{Misra2015,Rosenthal2015} but these delivered very poor results and are not reported in this article.
The full set of features and their parameters are described in the supplementary material to this article.
We experiment with the full feature set, individual feature groups, and feature ablation (all features except for one group).

\subsection{Deep Learning Approaches}
\label{sec:deep-learning}
As alternatives to our feature-based systems, we consider three deep learning approaches. The first is the Convolutional Neural Net of Kim \cite{Kim:2014}
which has shown to perform excellently on many diverse classification tasks such as sentiment analysis and question classification and is still a strong competitor 
among 
neural techniques focusing on sentence classification \cite{Komninos:2016,Zhang:Lee:2016,Zhang:2016}.  
We consider two variants of Kim's CNN, one in which words' vectors are initialized with pre-trained GoogleNews word embeddings
({CNN:w2vec}) and one in which the vectors are randomly initialized and updated during training ({CNN:rand}).
Our second model is an LSTM (long short-term memory) neural net for sentence classification (LSTM) and our third model is a bidirectional LSTM (BiLSTM).

For all neural network classifiers, we use default hyperparameters concerning hidden dimensionalities (for the two LSTM models), number of filters (for the convolutional neural net), and others.
We train each of the three neural networks for 15 iterations and choose in each case the learned model that performs best on a held-out development set of roughly 10\% of the training data as the model to apply to unseen test data. This corresponds to an early stopping regularization scheme.


\begin{table*}[ht]
\begin{small}
\begin{center}
\begin{tabular}{p{1.8cm}|rr|rr|rr|rr|rr|rr|rr} \toprule
Target $\rightarrow$ \newline System $\downarrow$ & \multicolumn{2}{c}{\textbf{MT}} & \multicolumn{2}{c}{\textbf{OC}} & \multicolumn{2}{c}{\textbf{PE}} & \multicolumn{2}{c}{\textbf{VG}} & \multicolumn{2}{c}{\textbf{WD}} & \multicolumn{2}{c}{\textbf{WTP}} & \multicolumn{2}{c}{\textbf{Average}} \\ \midrule
\multicolumn{15}{c}{\centering{\emph{neural network models}}} \\
BiLSTM & 68.8 & 41.8 & 58.0 & 22.4 & 73.0 & 62.0 & 60.9 & 37.7 & 60.0 & 24.5 & 57.9 & 28.5 & 63.1 & 36.1 \\
CNN:rand & 78.6 & 67.3 & \bf{60.5} & \bf{25.6} & \bf{73.6} & 61.1 & \bf{65.9} & \bf{45.0} & 61.1 & 25.8 & 58.6 & 28.9 & 66.4 & 42.3 \\
CNN:w2vec & 73.7 & 60.9 & 58.2 & 23.7 & 74.0 & 61.7 & 63.8 & 33.5 & 62.6 & \bf{28.9} & 57.3 & 24.3 & 64.9 & 38.8 \\
LSTM & 65.2 & 48.3 & 58.5 & 22.3 & 71.8 & 60.7 & 61.3 & 40.1 & 61.6 & 25.9 & 58.0 & 28.4 & 62.7 & 37.6 \\
\midrule
LR & \multicolumn{12}{c}{\centering{\emph{feature ablation and combination}}} \\
-Discourse & 73.0 & 60.8 & 59.9 & 22.9 & 70.6 & 60.6 & 62.5 & 42.6 & 63.7 & 23.2 & 59.7 & 30.2 & 64.9 & 40.0 \\
-Embeddings & 74.6 & 62.9 & 59.6 & 22.6 & 70.4 & 60.4 & 62.9 & 43.1 & 63.9 & 23.5 & 59.4 & 29.9 & 65.1 & 40.4 \\
-Lexical & 72.1 & 59.5 & 59.6 & 22.5 & 65.9 & 55.1 & 60.8 & 40.5 & 60.1 & 18.5 & 57.7 & 27.8 & 62.7 & 37.3 \\
-Structure & 74.4 & 62.6 & 60.0 & 23.0 & 70.4 & 60.4 & 62.0 & 41.8 & 64.2 & 23.4 & 59.5 & 30.0 & 65.1 & 40.2 \\
-Syntax & \bf{79.8} & \bf{70.3} & 59.8 & 22.9 & 72.1 & \bf{62.5} & 63.4 & 43.8 & \bf{65.1} & 25.5 & \bf{60.1} & \bf{30.5} & \bf{66.7} & \bf{42.6} \\
All Features & 74.4 & 62.7 & 59.9 & 22.9 & 70.6 & 60.6 & 62.5 & 42.6 & 63.8 & 23.3 & 59.7 & 30.2 & 65.1 & 40.4 \\
\midrule
LR &\multicolumn{12}{c}{\centering{\emph{single feature groups}}} \\
+Discourse & 70.0 & 56.7 & 49.4 & 13.8 & 50.1 & 41.7 & 49.6 & 30.6 & 57.6 & 14.9 & 49.5 & 18.4 & 54.4 & 29.3 \\
+Embeddings & 72.4 & 59.8 & 58.8 & 20.8 & 68.2 & 57.7 & 59.7 & 39.3 & 64.2 & 23.8 & 59.0 & 28.9 & 63.7 & 38.4 \\
+Lexical & 75.9 & 64.7 & 59.5 & 21.4 & 71.8 & 62.1 & 61.1 & 40.5 & 64.0 & 22.2 & 59.0 & 27.7 & 65.2 & 39.8 \\
+Structure & 57.1 & 42.0 & 56.5 & 20.0 & 54.2 & 39.5 & 55.4 & 33.3 & 48.4 & 9.0 & 55.4 & 25.2 & 54.5 & 28.2 \\
+Syntax & 66.7 & 52.5 & 58.1 & 21.0 & 64.1 & 52.9 & 60.7 & 40.4 & 57.6 & 15.5 & 57.0 & 27.0 & 60.7 & 34.9 \\
\midrule
\multicolumn{15}{c}{\centering{\emph{baselines}}} \\
Majority bsl & 42.9 & 0.0 & 48.0 & 0.0 & 41.3 & 0.0 & 44.5 & 0.0 & 48.6 & 0.0 & 46.7 & 0.0 & 45.3 & 0.0 \\
Random bsl & 50.7 & 33.2 & 49.9 & 13.5 & 50.8 & 38.0 & 50.4 & 28.8 & 51.6 & 10.8 & 48.9 & 18.8 & 50.4 & 23.9 \\
\bottomrule
\end{tabular}
\end{center}
\end{small}
\caption{In-domain experiments, best values per column are highlighted.  For each dataset (column head) we show two scores: \emph{Macro-$F_1$} score (left-hand column) and $F_1$ score for claims (right-hand column).}
\label{tab:results-cross-validation}
\end{table*}

\section{Results}
\label{sec:results}

In the following, we summarize the results of the various learners described above. 
Obtaining all results required heavy computation, e.g.\ the cross-validation experiments for feature-based systems took 56 days of computing.
We intentionally do not list the results of previous work on those datasets.
The scores are not comparable since we strictly work on sentence level (rather than e.g. clause level) and applied downsampling to the training data. 
All reported significance tests were conduced using two-tailed Wilcoxon Signed-Rank Test for matched pairs, i.e. paired scores of $F_1$ scores from two compared systems \cite{Japkowicz.Shah.2014}.

\subsection{In-Domain Experiments}

The performance of the learners is quite divergent across datasets, with Macro-$F_1$ scores\footnote{Described as $\textrm{\emph{Fscore}}_M$ in \newcite{Sokolova.Lapalme.2009}.} ranging from 60\% (WTP) to 80\% (MT), average 67\% (see Table~\ref{tab:results-cross-validation}).
On all datasets, our best systems clearly outperform both baselines.
In isolation, lexical, embedding, and syntax features are most helpful, whereas structural features did not help in most cases. 
Discourse features only contribute significantly on MT. 
When looking at the performance of the feature-based approaches, the most striking finding is the importance of lexical (in our setup, unigram) information.

The average performances of LR$_{-syntax}$ and CNN:rand are virtually identical, both for Macro-$F_1$ and Claim-$F_1$, with a slight advantage for the feature-based approach,  but their difference is not statistically significant ($p \leq 0.05$).
Altogether, these two systems exhibit significantly better average performances than all other models surveyed here, both those relying on 
and those not relying on hand-crafted features ($p \leq 0.05$).
The absence or the different nature of inter-annotator agreement measures for all datasets prevent us from searching for correlations between agreement and performance. 
But we observed that the systems yield better results on PE and MT, both datasets with good inter-annotator agreement ($\alpha_{u}$ = 0.77 for PE and $\kappa$ = 0.83 for MT).

\begin{table*}[ht]
\begin{small}
\begin{center}
\begin{tabular}{p{1.8cm}|rr|rr|rr|rr|rr|rr|rr} \toprule
Target $\rightarrow$ \newline Source/Sys. $\downarrow$ & \multicolumn{2}{c}{\textbf{MT}} & \multicolumn{2}{c}{\textbf{OC}} & \multicolumn{2}{c}{\textbf{PE}} & \multicolumn{2}{c}{\textbf{VG}} & \multicolumn{2}{c}{\textbf{WD}} & \multicolumn{2}{c}{\textbf{WTP}} & \multicolumn{2}{c}{\textbf{Average}} \\ \midrule
& \multicolumn{12}{c|}{\centering{\emph{CNN:rand}}} &&  \\
MT & \it{78.6} & \it{67.3} & 51.0 & 7.4 & 56.9 & 22.1 & 57.2 & 15.7 & 52.4 & 9.4 & 49.4 & 10.9 & 53.4 & 13.1 \\
OC & 57.1 & 39.7 & \it{60.5} & \it{25.6} & 56.4 & 42.8 & 58.9 & 37.3 & 54.6 & 13.2 & \bf{58.4} & \bf{28.9} & 57.1 & 32.4 \\
PE & 59.8 & 18.0 & 54.2 & 9.5 & \it{73.6} & \it{61.1} & 57.5 & 18.7 & \bf{55.5} & \bf{15.9} & 54.7 & 16.0 & 56.3 & 15.6 \\
VG & \bf{68.7} & \bf{51.5} & 55.8 & 19.2 & \bf{57.0} & 32.0 & \it{65.9} & \it{45.0} & 51.7 & 10.5 & 54.7 & 22.0 & 57.6 & 27.0 \\
WD & 64.4 & 3.5 & 51.3 & 1.3 & 41.3 & 0.0 & 44.5 & 0.0 & \it{61.1} & \it{25.8} & 46.7 & 0.0 & 49.6 & 1.0 \\
WTP & 58.5 & 26.6 & 56.8 & 15.4 & 56.0 & 18.5 & 55.3 & 19.4 & 52.9 & 11.6 & \it{58.6} & \it{28.9} & 55.9 & 18.3 \\
\it{Average} & 61.7 & 27.9 & 53.8 & 10.6 & 53.5 & 23.1 & 54.7 & 18.2 & 53.4 & 12.1 & 52.8 & 15.6 & 55.0 & 17.9 \\
\midrule
& \multicolumn{12}{c|}{\centering{\emph{LR All features}}} &  &  \\
MT & \it{74.4} & \it{62.7} & 53.9 & 17.0 & 51.9 & 29.5 & 56.1 & 34.2 & 55.1 & 14.5 & 52.5 & 21.2 & 53.9 & 23.3 \\
OC & 60.0 & 45.1 & \it{59.9} & \it{22.9} & 56.7 & \bf{47.0} & 58.6 & \bf{38.0} & 54.1 & 12.2 & 57.7 & 27.5 & 57.4 & \bf{34.0} \\
PE & 58.1 & 36.3 & 54.6 & 17.3 & \it{70.6} & \it{60.6} & 54.1 & 21.4 & 54.0 & 13.5 & 54.4 & 20.4 & 55.0 & 21.8 \\
VG & 65.8 & 51.4 & \bf{57.3} & \bf{21.7} & \bf{57.0} & 45.1 & \it{62.5} & \it{42.6} & 54.5 & 13.1 & 55.1 & 24.8 & \bf{57.9} & 31.2 \\
WD & 62.6 & 38.5 & 55.4 & 19.0 & 56.0 & 30.1 & 55.1 & 23.3 & \it{63.8} & \it{23.3} & 53.6 & 20.9 & 56.5 & 26.3 \\
WTP & 58.0 & 41.7 & 56.1 & 20.3 & 56.8 & 42.6 & \bf{59.1} & \bf{38.0} & 52.2 & 11.2 & \it{59.7} & \it{30.2} & 56.5 & 30.8 \\
\it{Average} & 60.9 & 42.6 & 55.5 & 19.1 & 55.7 & 38.9 & 56.6 & 31.0 & 54.0 & 12.9 & 54.7 & 23.0 & 56.2 & 27.9 \\
\midrule
LR& \multicolumn{12}{c|}{\centering{\emph{single feature groups (averages across all source domains)}}} &  &  \\
+Discourse & 40.2 & 15.0 & 31.7 & 5.8 & 30.3 & 27.4 & 27.7 & 19.9 & 40.9 & 4.5 & 25.3 & 13.3 & 32.7 & 14.3 \\
+Embeddings & 56.6 & 35.2 & 51.4 & 12.8 & 53.6 & 30.7 & 53.3 & 24.3 & 54.2 & 13.2 & 52.9 & 19.0 & 53.7 & 22.5 \\
+Lexical & 61.0 & 42.2 & 55.2 & 18.3 & 56.2 & 38.6 & 54.7 & 29.1 & 53.1 & 11.9 & 54.9 & 23.4 & 55.9 & 27.2 \\
+Structure  & 44.2 & 22.9 & 53.6 & 18.5 & 52.5 & 38.4 & 53.6 & 32.1 & 49.1 & 9.0 & 53.4 & 23.3 & 51.1 & 24.0 \\
+Syntax & 54.8 & 37.0 & 54.2 & 17.5 & 54.3 & 40.6 & 55.7 & 32.0 & 53.0 & 11.8 & 53.8 & 22.5 & 54.3 & 26.9 \\
\midrule

& \multicolumn{12}{c|}{\centering{\emph{baselines}}} &  &  \\
Majority bsl & 42.9 & 0.0 & 48.0 & 0.0 & 41.3 & 0.0 & 44.5 & 0.0 & 48.6 & 0.0 & 46.7 & 0.0 & 45.3 & 0.0 \\
Random bsl & 47.5 & 30.6 & 50.5 & 14.0 & 51.0 & 38.4 & 51.0 & 29.3 & 49.3 & 9.3 & 50.3 & 20.2 & 49.9 & 23.6 \\
\bottomrule
\end{tabular}
\end{center}
\end{small}
\caption{Cross-domain experiments, best values per column are highlighted, in-domain results (for comparison) in italics; results only for selected systems. For each source/target combination we show two scores: \emph{Macro-$F_1$} score (left-hand column) and $F_1$ score for claims (right-hand column).}
\label{tab:results-cross-domain-SINGLE}
\end{table*}

\subsection{Cross-Domain Experiments}
\label{sect:Cross-Domain-Experiments}

For all six datasets, training on different sources resulted in a performance drop.
Table~\ref{tab:results-cross-domain-SINGLE} lists the results of the best feature-based (\emph{LR All features}) and deep learning (\emph{CNN:rand}) systems, as well as single feature groups (averages over all source domains, results for individual source domains can be found in the supplementary material to this article).
We note the biggest performance drops on the datasets which performed best in the in-domain setting (MT and PE).
For the lowest scoring datasets, OC and WTP, the differences are only marginal when trained on a suitable dataset (VG and OC, respectively).
The best feature-based approach outperforms the best deep learning approach in most scenarios.
In particular, as opposed to the in-domain experiments, the difference of the Claim-$F_1$ measure between the feature-based approaches and the deep learning approaches is striking. 
In the feature-based approaches, on average, a combination of all features yields the best results for both Macro-$F_1$ and Claim-$F_1$.
When comparing single features, lexical ones do the best job.

Looking at the best overall system (LR with all features), the average test results when training on different source datasets are between 54\% Macro-$F_1$ resp.\ 23\% Claim-$F_1$ (both MT) and 58\% (VG) resp.\ 34\% (OC).
Depending on the goal that should be achieved, training on VG (highest average Macro-$F_1$) or OC (highest average Claim-$F_1$) seems to be the best choice when the domain of test data is unknown (we analyze this finding in more depth in §\ref{sec:analysis}).
MT clearly gives the best results as target domain, followed by PE and VG. 

We also performed experiments with mixed sources, the results are shown in Table~\ref{tab:results-LODO}.
We did this in a leave-one-domain-out fashion, in particular we trained on all but one datasets and tested on the remaining one.
In this scenario, the neural network systems seem to benefit from the increased amount of training data and thus gave the best results.
Overall, the mixed sources approach works better than many of the single-source cross-domain systems -- yet, the differences were not found to be significant, but as good as training on suitable single sources (see above).

\begin{table*}[ht]
\begin{small}
\begin{center}
\begin{tabular}{p{1.8cm}|rr|rr|rr|rr|rr|rr|rr} \toprule
Target $\rightarrow$ \newline System $\downarrow$ & \multicolumn{2}{c}{\textbf{MT}} & \multicolumn{2}{c}{\textbf{OC}} & \multicolumn{2}{c}{\textbf{PE}} & \multicolumn{2}{c}{\textbf{VG}} & \multicolumn{2}{c}{\textbf{WD}} & \multicolumn{2}{c}{\textbf{WTP}} & \multicolumn{2}{c}{\textbf{Avg}} \\ \toprule
CNN:rand & 62.8 & 41.4 & \textbf{57.8} & \textbf{22.4} & \textbf{59.7} & 36.2 & \textbf{58.6} & 28.1 & \textbf{54.2} & \textbf{14.1} & \textbf{56.8} & 25.6 & \textbf{58.3} & 28.0 \\
\midrule
All features & \textbf{64.7} & \textbf{49.5} & 56.4 & 20.6 & 57.8 & \textbf{45.8} & 58.2 & \textbf{36.4} & 52.3 & 11.3 & 56.0 & \textbf{26.0} & 57.6 & \textbf{31.6} \\

\midrule\midrule
Majority bsl & 42.9 & 0.0 & 48.0 & 0.0 & 41.3 & 0.0 & 44.5 & 0.0 & 48.6 & 0.0 & 46.7 & 0.0 & 45.3 & 0.0 \\
Random bsl & 47.5 & 30.6 & 50.5 & 14.0 & 51.0 & 38.4 & 51.0 & 29.3 & 49.3 & 9.3 & 50.3 & 20.2 & 49.9 & 23.6 \\
\bottomrule
\end{tabular}
\end{center}
\end{small}
\caption{Leave-one-domain-out experiments, best values per column are highlighted. For each test dataset (column head) we show two scores: \emph{Macro-$F_1$} score (left-hand column) and $F_1$ score for claims (right-hand column).}
\label{tab:results-LODO}
\end{table*}


\section{Further Analysis and Discussion}
\label{sec:analysis}

To better understand which factors influence 
cross-domain performance of the systems we tested, we considered the following variables 
as potential determinants of 
outcome: 
similarity between source and target domain,
the source domain itself,
training data size, and
the ratio between claims and non-claims.

\begin{table}
\begin{small}
\pgfplotstabletypeset[colorcells]{
x,MT,OC,PE,VG,WD,WTP
MT,100,47,51,52,49,48
OC,56,100,55,68,71,71
PE,59,58,100,66,67,57
VG,51,58,52,100,59,62
WD,54,61,61,62,100,55
WTP,49,59,49,57,57,100
}
\end{small}
\caption{\label{tab:correlations} Heatmap of Spearman correlations in \% based on most frequent 500 lemmas for each dataset. Source domain: rows, target domain: columns.}
\end{table}

We calculated the Spearman correlation of the top-500 lemmas between the datasets in each direction, see results in Table~\ref{tab:correlations}.
The most similar domains are OC (source $s$) and WTP (target $t$), coming from the same authors.  
OC ($s$) and WD ($t$) as well OC ($s$) and VG ($t$) are also highly correlated. 
For a statistical test of potential correlations between cross-domain performances and the introduced variables, 
we regress the cross-domain results (Table~\ref{tab:results-cross-domain-SINGLE}) on Table~\ref{tab:correlations} ($\text{T5}$ in the following equation),
on the number of claims $\#\text{C}$ (directly related to training data size in our experiments, effect of downsampling), and
on the ratio of claims to non-claims $\text{R}$.\footnote{Overall, we had 15 different systems, see upper 15 rows in Table~\ref{tab:results-cross-validation}. Therefore, we had 15 different regression models.}
More precisely, given source/training data and target data pairs $(s,t)$ in Table~\ref{tab:results-cross-domain-SINGLE}, we estimate the linear regression model
\begin{equation}
  y_{st} = \alpha\cdot\text{T5}_{st}
  +\beta\cdot\log(\#\text{C}_s)+\gamma\cdot\text{R}_t+\epsilon_{st},
\end{equation}
where $y_{st}$ denotes the Macro-$F_1$ score when training on $s$ and testing on $t$. In the regression, we also include binary dummy variables $\mathbbm{1}_{\sigma}=\mathbbm{1}_{s,\sigma}$ for each domain $\sigma$ whose value is $1$ if 
$s=\sigma$ (and 0 otherwise).
These help us identify ``good'' source domains.

The coefficient $\alpha$ for Table~\ref{tab:correlations} is not statistically significantly different from zero in any case. 
Ultimately, this means that it is difficult to predict cross-domain performance from lexical similarity of the datasets. 
This is in contrast to e.g., POS tagging, where lexical similarity has been reported to predict cross-domain performance very well \cite{VanAsch:2010}.
The coefficient for training data size $\beta$ is statistically significantly different from zero in three out of 15 cases.
In particular, it is significantly positive in two (CNN:rand, CNN:w2vec) out of four cases for the neural networks. 
This indicates that the neural networks would have particularly benefited from more training data, which is confirmed by the improved performance of the neural networks in the mixed sources experiments (cf. §\ref{sect:Cross-Domain-Experiments}). 
The ratio of claims to non-claims in $t$ is 
among the best predictors for the variables considered here (coefficient $\gamma$ is significant in three out of 15 cases, but consistently positive).
This is probably due to our decision to balance training data (downsampling non-claims) to keep the assessment of claim identification realistic for real-world applications, where the class ratio of $t$ is unknown.
Our systems are thus inherently biased towards a higher claim ratio.

Finally, the dummy variables for OC and VG are three times significantly positive, but consistently positive overall. 
Their average coefficient is 2.31 and 1.90, respectively, while the average coefficients for all other source datasets is negative, and not significant in most cases. 
Thus, even when controlling for all other factors such as training data size and the different claim ratios of target domains, OC and VG are the best source domains for cross-domain claim classification in our experiments. 
OC and VG are particularly good training sources for the detection of claims (as opposed to non-claims)---the minority class in all datasets---as indicated by the average Claim-$F_1$ scores in Table~\ref{tab:results-cross-domain-SINGLE}. 

One finding that was confirmed both in-domain as well as cross-domain was the importance of lexical features as compared to other feature groups.
As mere lexical similarity between domains does not explain performance (cf. coefficient $\alpha$ above), this finding indicated that the learners relied on a few, but important lexical clues. 
To go more into depth, we carried out error analysis on the CNN:rand cross-domain results.
We used OC, VG and PE as source domains, and MT and WTP as target domains.
By examining examples in which a model trained on OC and VG made correct predictions as opposed to a model trained on PE, we quickly noticed that lexical indicators indeed played a crucial role.
In particular, the occurrence of the word ``should'' (and to a lower degree: ``would'', ``article'', ``one'') are helpful for the detection of claims across various datasets.
In MT, a simple baseline labeling every sentence containing ``should'' as claim achieves 76.1 Macro-$F_1$ (just slightly below the best in-domain system on this dataset).
In the other datasets, this phenomenon is far less dominant, but still observable.
We conclude that a few rather simple rules (learned by models trained on OC and VG, but not by potentially more complex models trained on PE) make a big difference in the cross-domain setting. 

\section{Conclusion}
In a rigorous empirical assessment of different machine learning systems, we compared how six datasets model claims as the fundamental component of an argument.
The varying performance of the tested in-domain systems reflects different notions of claims also observed in a qualitative study of claims across the domains.
Our results reveal that the best in-domain system is not necessarily the best system in environments where the target domain is unknown. 
Particularly, we found that mixing source domains and training on two rather noisy datasets (OC and VG) gave the best results in the cross-domain setup.
The reason for this seem to be a few important lexical indicators (like the word ``should'') which are learned easier under these circumstances.
In summary, as for the six datasets we analyzed here, our analysis shows that the essence of a claim is not much more than a few lexical clues. 

From this, we conclude that future work should address the problem of vague conceptualization of claims as central components of arguments. 
A more consistent notion of claims, which also holds across domains, would potentially not just benefit cross-domain claim identification, but also higher-level applications relying on argumentation mining \cite{wachsmuth-stein-ajjour:2017:EACLlong}.
To further overcome the problem of domain dependence, multi-task learning is a framework that could be explored \cite{Sogaard2016} for different conceptualizations of claims.


\section*{Acknowledgments}
This work has been supported by the German Federal Ministry of Education and Research (BMBF) under the promotional reference 03VP02540 (ArgumenText), by the GRK 1994/1 AIPHES (DFG), and by the ArguAna Project GU~798/20-1 (DFG).

\bibliography{emnlp2017}
\bibliographystyle{emnlp_natbib}
\clearpage

\section*{Supplementary Material}
\input{supplementary-material}

 \end{document}

%% file: supplementary-material.tex
\subsection*{Experimental Setup: Detailed Description of Hand-Crafted Features}

This part of the supplementary material describes the hand-crafted features we used in more detail.

\noindent
\textbf{Structure Features}: 
Structure features capture the position, the length and the punctuation of a sentence. 
First, we define two binary features which indicate if the current sentence is the first or last sentence in the paragraph in which it is contained. 
These feature are motivated by the findings of \newcite{Stab2016} who found that structural properties of argument components are effective for distinguishing the argumentative function of argument components. 
In addition, \newcite{Peldszus2016} found that $43\%$ of claims appear in the first sentence. 
Second, we add the number of tokens of the sentence to our feature set which proved  to be indicative for identifying argumentatively relevant sentences \cite{Biran2011a,Moens2007}.
Finally, we adopted the punctuation features from \newcite{MochalesPalau2009}.

\noindent
\textbf{Lexical Features}: 
As lexical features, we employ lowercased unigrams. 
We assume that these features are helpful for detecting claims since they capture discourse connectors like ``\emph{therefore}'', ``\emph{thus}'', or ``\emph{hence}'' which frequently signal the presence of claims. 
We consider the most frequent 4,000 unigrams as binary features.

\noindent
\textbf{Syntactic Features}:
To account for grammatical information at the sentence level, we include information about the part-of-speech and parse tree for each sentence.
Following \newcite{Stab2016}, we add binary POS $n$-grams (the 2000 most frequent, $2 \leq n \leq 4$) and constituent parse tree production rules (4000 most frequent, minimum occurrence 5) as originally suggested by \newcite{Lin2009}.
Additionally, to account for the frequency of POS tags, we include a feature which counts the occurrence of each part-of-speech per sentence.

\noindent
\textbf{Discourse Features}:
\newcite{Cabrio2013} suggested that the relation between parts of discourse (e.g. connectives such as ``because'') can be helpful to determine argumentative content.
As this finding is affirmed by \newcite{Stab2016}, we include discourse features with the help of the Penn Discourse Treebank (PDTB) styled end-to-end discourse parser as presented by \newcite{Lin2014}.
We include discourse relations extracted from the parser output as a triple of i) the type of relation, ii) whether the relation is implicit or explicit, and iii) whether the current sentence is part of the first or the second discourse argument (or both).

\noindent
\textbf{Embedding Features}:
We represent each sentence as a summation of its word embeddings \cite{Guo.et.al.2014}. 
These simple yet powerful latent semantic representation features have been found predictive in related work \cite{Habernal2015,Habernal2016}. 
In particular, we use 
pre-trained 300 dimensional GoogleNews word embeddings.\footnote{\url{https://code.google.com/archive/p/word2vec/}}

\subsection*{Cross-Domain Experiments: Full Results}

Table~\ref{tab:results-cross-domain-apdx} displays the results of cross-domain experiments, for all source domains. 
Precisely, we list results for the six best in-domain systems, according to average F$_1$ scores.

\begin{table*}[ht]
\begin{small}
\begin{center}
\begin{tabular}{p{2cm}|rr|rr|rr|rr|rr|rr|rr} \toprule
Train $\downarrow$ Test $\rightarrow$ & \multicolumn{2}{c}{\textbf{MT}} & \multicolumn{2}{c}{\textbf{OC}} & \multicolumn{2}{c}{\textbf{PE}} & \multicolumn{2}{c}{\textbf{VG}} & \multicolumn{2}{c}{\textbf{WD}} & \multicolumn{2}{c}{\textbf{WTP}} & \multicolumn{2}{c}{\textbf{Avg}} \\ \midrule
& \multicolumn{12}{c|}{\centering{\emph{LR+Lexical}}} & \textit{55.9} & \textit{27.2} \\
MT & -- & -- & 53.6 & 16.5 & 55.3 & 28.3 & 53.4 & 25.3 & 54.8 & 14.5 & 53.3 & 20.8 & 54.1 & 21.1 \\
OC & 62.3 & 46.4 & -- & -- & 57.3 & 48.0 & 59.8 & 39.2 & 54.2 & 12.0 & 57.3 & 26.5 & 58.2 & 34.4 \\
PE & 61.5 & 46.5 & 54.6 & 17.8 & -- & -- & 53.8 & 32.9 & 53.1 & 11.5 & 54.7 & 24.2 & 55.5 & 26.6 \\
VG & 62.7 & 47.0 & 57.4 & 20.4 & 57.2 & 47.4 & -- & -- & 50.2 & 10.0 & 55.6 & 25.4 & 56.6 & 30.0 \\
WD & 56.7 & 24.1 & 54.3 & 17.5 & 55.0 & 22.8 & 51.4 & 12.7 & -- & -- & 53.5 & 20.2 & 54.2 & 19.4 \\
WTP & 61.8 & 46.8 & 56.3 & 19.1 & 56.4 & 46.4 & 55.3 & 35.2 & 53.2 & 11.7 & -- & -- & 56.6 & 31.9 \\
\midrule
&\multicolumn{12}{c|}{\centering{\emph{LR-Embeddings}}} & \textit{56.1} & \textit{28.0} \\
MT & -- & -- & 54.3 & 17.4 & 52.0 & 30.0 & 56.3 & 34.5 & 55.1 & 14.5 & 52.6 & 21.4 & 54.0 & 23.5 \\
OC & 58.4 & 43.8 & -- & -- & 56.7 & 46.9 & 59.0 & 38.4 & 54.3 & 12.4 & 57.3 & 27.2 & 57.1 & 33.7 \\
PE & 58.6 & 37.0 & 55.0 & 18.2 & -- & -- & 53.8 & 20.9 & 53.6 & 13.0 & 54.5 & 21.0 & 55.1 & 22.0 \\
VG & 64.5 & 49.8 & 57.1 & 21.6 & 57.0 & 45.2 & -- & -- & 54.3 & 13.0 & 55.3 & 25.1 & 57.7 & 31.0 \\
WD & 63.3 & 41.5 & 55.7 & 19.5 & 55.9 & 31.5 & 55.0 & 23.6 & -- & -- & 53.7 & 21.2 & 56.7 & 27.5 \\
WTP & 57.7 & 41.6 & 56.0 & 19.9 & 56.2 & 42.5 & 57.2 & 35.8 & 52.8 & 11.6 & -- & -- & 56.0 & 30.3 \\
\midrule
&\multicolumn{12}{c|}{\centering{\emph{LR-Structure}}} & \textit{56.0} & \textit{27.8} \\
MT & -- & -- & 52.7 & 15.6 & 51.3 & 29.9 & 56.2 & 34.7 & 55.6 & 15.0 & 51.7 & 20.6 & 53.5 & 23.2 \\
OC & 59.0 & 44.2 & -- & -- & 56.3 & 46.7 & 58.8 & 38.3 & 54.2 & 12.3 & 57.7 & 27.5 & 57.2 & 33.8 \\
PE & 57.5 & 35.3 & 54.8 & 17.7 & -- & -- & 54.0 & 21.1 & 53.7 & 13.2 & 54.3 & 20.3 & 54.9 & 21.5 \\
VG & 65.6 & 51.3 & 57.0 & 21.3 & 56.8 & 44.9 & -- & -- & 54.5 & 13.2 & 55.1 & 24.8 & 57.8 & 31.1 \\
WD & 62.8 & 39.1 & 55.5 & 19.2 & 55.6 & 29.8 & 55.3 & 24.3 & -- & -- & 53.5 & 21.0 & 56.5 & 26.7 \\
WTP & 58.2 & 41.8 & 56.1 & 20.2 & 56.7 & 42.5 & 57.8 & 36.5 & 52.7 & 11.6 & -- & -- & 56.3 & 30.5 \\
\midrule
&\multicolumn{12}{c|}{\centering{\emph{LR-Syntax}}} & \textit{56.2} & \textit{25.9} \\
MT & -- & -- & 53.4 & 16.3 & 55.2 & 29.0 & 55.3 & 28.4 & 55.1 & 14.9 & 53.1 & 21.0 & 54.4 & 21.9 \\
OC & 63.8 & 48.7 & -- & -- & 57.9 & 47.8 & 59.1 & 38.5 & 54.1 & 12.2 & 57.3 & 27.2 & 58.4 & 34.9 \\
PE & 60.7 & 40.7 & 53.5 & 9.0 & -- & -- & 55.7 & 24.6 & 53.1 & 12.3 & 53.1 & 13.6 & 55.2 & 20.0 \\
VG & 67.3 & 53.2 & 56.9 & 21.1 & 58.2 & 45.6 & -- & -- & 51.9 & 11.0 & 55.8 & 25.7 & 58.0 & 31.3 \\
WD & 57.9 & 19.9 & 53.9 & 16.9 & 55.6 & 19.7 & 51.6 & 9.9 & -- & -- & 53.2 & 18.9 & 54.5 & 17.1 \\
WTP & 62.5 & 45.7 & 56.0 & 20.0 & 56.8 & 39.4 & 56.0 & 34.4 & 53.2 & 12.3 & -- & -- & 56.9 & 30.3 \\
\midrule
& \multicolumn{12}{c|}{\centering{\emph{LR All features}}} & \textit{56.2} & \textit{27.9} \\
MT & -- & -- & 53.9 & 17.0 & 51.9 & 29.5 & 56.1 & 34.2 & 55.1 & 14.5 & 52.5 & 21.2 & 53.9 & 23.3 \\
OC & 60.0 & 45.1 & -- & -- & 56.7 & 47.0 & 58.6 & 38.0 & 54.1 & 12.2 & 57.7 & 27.5 & 57.4 & 34.0 \\
PE & 58.1 & 36.3 & 54.6 & 17.3 & -- & -- & 54.1 & 21.4 & 54.0 & 13.5 & 54.4 & 20.4 & 55.0 & 21.8 \\
VG & 65.8 & 51.4 & 57.3 & 21.7 & 57.0 & 45.1 & -- & -- & 54.5 & 13.1 & 55.1 & 24.8 & 57.9 & 31.2 \\
WD & 62.6 & 38.5 & 55.4 & 19.0 & 56.0 & 30.1 & 55.1 & 23.3 & -- & -- & 53.6 & 20.9 & 56.5 & 26.3 \\
WTP & 58.0 & 41.7 & 56.1 & 20.3 & 56.8 & 42.6 & 59.1 & 38.0 & 52.2 & 11.2 & -- & -- & 56.5 & 30.8 \\
\midrule
& \multicolumn{12}{c|}{\centering{\emph{CNN:rand}}} & \textit{55.0} & \textit{17.9} \\
MT & -- & -- & 51.0 & 7.4 & 56.9 & 22.1 & 57.2 & 15.7 & 52.4 & 9.4 & 49.4 & 10.9 & 53.4 & 13.1 \\
OC & 57.1 & 39.7 & -- & -- & 56.4 & 42.8 & 58.9 & 37.3 & 54.6 & 13.2 & 58.4 & 28.9 & 57.1 & 32.4 \\
PE & 59.8 & 18.0 & 54.2 & 9.5 & -- & -- & 57.5 & 18.7 & 55.5 & 15.9 & 54.7 & 16.0 & 56.3 & 15.6 \\
VG & 68.7 & 51.5 & 55.8 & 19.2 & 57.0 & 32.0 & -- & -- & 51.7 & 10.5 & 54.7 & 22.0 & 57.6 & 27.0 \\
WD & 64.4 & 3.5 & 51.3 & 1.3 & 41.3 & 0.0 & 44.5 & 0.0 & -- & -- & 46.7 & 0.0 & 49.6 & 1.0 \\
WTP & 58.5 & 26.6 & 56.8 & 15.4 & 56.0 & 18.5 & 55.3 & 19.4 & 52.9 & 11.6 & -- & -- & 55.9 & 18.3 \\

\midrule\midrule
Majority bsl & 42.9 & 0.0 & 48.0 & 0.0 & 41.3 & 0.0 & 44.5 & 0.0 & 48.6 & 0.0 & 46.7 & 0.0 & 45.3 & 0.0 \\
Random bsl & 47.5 & 30.6 & 50.5 & 14.0 & 51.0 & 38.4 & 51.0 & 29.3 & 49.3 & 9.3 & 50.3 & 20.2 & 49.9 & 23.6 \\
\bottomrule
\end{tabular}
\end{center}
\end{small}
\caption{Cross-domain experiments, results only for selected systems. For each test dataset (column head) we show two scores: \emph{Macro $F_1$} score (left-hand column) and $F_1$ score for claims (right-hand column).}
\label{tab:results-cross-domain-apdx}
\end{table*}